\documentclass[11pt,a4paper]{article}

\usepackage{a4wide}
\usepackage{multirow}
\usepackage{empheq,fancybox}
\usepackage{graphicx}
\usepackage{amsfonts}
\usepackage{subfig}
\usepackage[figuresright]{rotating}
\usepackage{tabularx}
\usepackage[table]{xcolor}
\usepackage{pstricks,pstricks-add,pst-node,pst-plot,pst-coil}
\usepackage{algorithm}
\usepackage{algorithmic}
\usepackage{url}

\newcommand{\drate}{d_{\mathrm{rate}}}
\newcommand{\lsize}{l_{\mathrm{size}}}

\newcommand{\ccg}{\cellcolor{lightgray}}

\begin{document}

\title{A Hybrid Evolutionary Algorithm Based on Solution Merging for the Longest Arc-Preserving Common Subsequence Problem}

\author{Christian Blum$^{1}$ and Maria J.~Blesa$^{2}$  \\
~\\
$^1$Artificial Intelligence Research Institute (IIIA-CSIC)\\ 
Campus UAB, Bellaterra, Spain\\
{\sf christian.blum@iiia.csic.es}\\
~\\
$^2$Computer Science Deptartment\\
Universitat Polit{\`e}cnica de Catalunya (BarcelonaTech), Barcelona, Spain\\
{\sf mjblesa@cs.upc.es}}

\date{}

\maketitle

\begin{abstract}
The longest arc-preserving common subsequence problem is an NP-hard combinatorial optimization problem from the field of computational biology. This problem finds applications, in particular, in the comparison of arc-annotated Ribonucleic acid (RNA) sequences. In this work we propose a simple, hybrid evolutionary algorithm to tackle this problem. The most important feature of this algorithm concerns a crossover operator based on solution merging. In solution merging, two or more solutions to the problem are merged, and an exact technique is used to find the best solution within this union. It is experimentally shown that the proposed algorithm outperforms a heuristic from the literature. 
\end{abstract}

\section{Introduction}

\begin{figure*}[!t]
\centering
\subfloat[\label{fig:example:a}]{
  \includegraphics[width=0.21\textwidth]{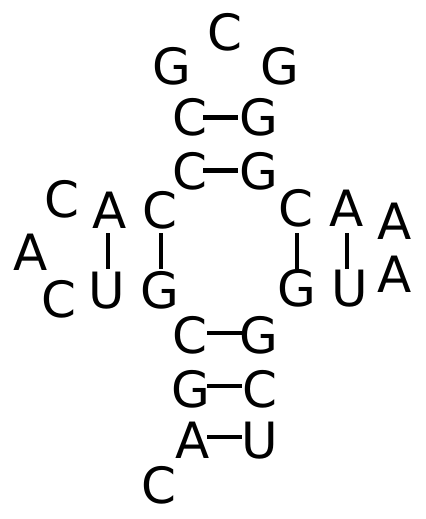}
} 
\subfloat[\label{fig:example:b}]{
  \includegraphics[width=0.48\textwidth]{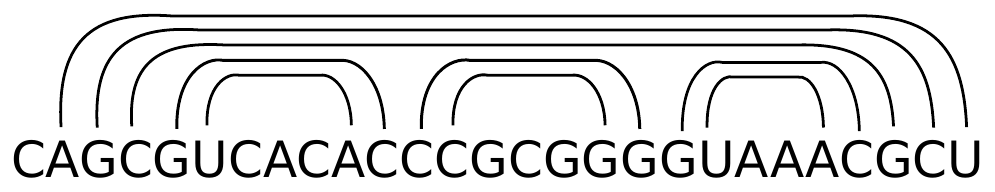}
}
\caption{(a) Example of the secondary structure of an RNA molecule. (b) The corresponding arc-annotated sequence. The example is reproduced from~\cite{Jiang2004257}.}
\label{fig:example}
\end{figure*}

In computer science, a \emph{string} (or sequence) $x$ of length $l_x$ is defined as a finite sequence of characters from a finite alphabet $\Sigma$. A string is a data type used to represent and store information. Words in a specific language, for example, are stored in a computer in terms of strings. Even whole texts may be stored by means of strings. Apart from fields such as information and text processing, strings arise, in particular, in the field of computational biology. This is because most of the genetic instructions involved in the growth, development, functioning and reproduction of living organisms are stored in \emph{Deoxyribonucleic acid} (DNA) and \emph{Ribonucleic acid} (RNA) molecules, which are double-stranded (in the case of DNA) or single-stranded (in the case of RNA) sequences of nucleotides. Hereby, each nucleotide is composed of a nitrogenous base, a five-carbon sugar (ribose or deoxyribose), and at least one phosphate group. Nucleotides in the context of RNA have one of four different nitrogenous bases: guanine (G), uracil (U), adenine (A), and cytosine (C). Therefore, any RNA molecule can be represented as a string of symbols from $\Sigma = \{G,U,A,C\}$. Such a string is called the \emph{primary structure} of an RNA molecule. However, RNA molecules generally fold in space, and different nucleotides bind together, for example, by means of hydrogene bonds. In simplified terms, G can only bind with C and U can only bind with A. En example of the secondary structure of an RNA molecule is shown in Figure~\ref{fig:example:a}.

In computer science terms, the hydrogene bonds of the secondary structure of an RNA sequence $x$ can be represented by a so-called \emph{arc annotation set} $P_x$, which is an unordered set of pairs of positions of $x$. Henceforth, the positions of a string $x$ range from 1 to $l_x$. Each pair $(i_1, i_2) \in P_x$ is called an \emph{arc annotation} (or simply an arc) between positions $i_1$ and $i_2$. As a convention, it must hold that $i_1 < i_2$. Moreover, $i_1$ is called the \emph{left endpoint} of arc $(i_1, i_2)$, and $i_2$ is called the \emph{right endpoint}. A pair $(x, P_x)$ is called an \emph{arc-annotated sequence}~\cite{Evans1999}. Note that the secondary structure of an RNA sequence can be described by an arc-annotated sequence. For an example see Figure~\ref{fig:example:b}. In fact, arc-annotated sequences have been widely used for this purpose (see, for example,~\cite{chiu2016comprehensive}). In particular, arc-annotated sequences have been useful in the context of the structural comparison of RNA sequences. An important way of comparing two (or more) sequences consists in computing their \emph{longest common subsequence} (LCS). Given a sequence $x$ over a finite alphabet $\Sigma$, sequence $t$ is called a \emph{subsequence} of $x$, if $t$ can be produced from $x$ by deleting characters. Given a set of input strings $\{s_1, \ldots, s_n\}$, the problem of finding the longest commons subsequence of all input strings is, in general, NP-hard~\cite{Mai78:jacm}. The best techniques available nowadays for solving this problem are based on beam search~\cite{Blum20093178}.

\subsection{Contribution of this Work}

The longest common subsequence problem in the context of arc-annotated sequences---the \emph{longest arc-preserving common subsequence} (LAPCS) problem---has first been introduced in~\cite{evans1999algorithms,Evans1999}. In particular, in the same works it was shown that the most general case of the problem (without any restrictions on the arcs) is NP-hard. In the meanwhile, five different variants of the problem---that is, restrictions of the general problem---have been studied in the related literature, and efficient algorithms were developed for three of these variants.\footnote{More details are given in Section~\ref{sec:lapcs}.} However, as far as we know, only one algorithm that is applicable to the most general case has been proposed so far (see~\cite{Jiang2004257}). In this work, we first phrase the LAPCS problem in the form of an integer linear program (ILP)~\cite{nemhauser:88}. Then we make use of this ILP in the context of a simple evolutionary algorithm based on \emph{solution merging}, where two or more solutions are merged and the best solution in this union is derived by means of an exact technique. Aggarwal et al.~\cite{aggarwal:97} originally suggested such an approach, labeled \emph{optimized crossover}, for the independent set problem. We provide an extensive experimental comparison of the proposed algorithm in comparison to the heuristic from the literature and to the application of a randomized multi-start heuristic. 

\subsection{Outline of the Paper}

The remainder of this paper is structured as follows. In Section~\ref{sec:lapcs}, a technical description of the tackled problem is provided. In subsequent sections---see Section~\ref{sec:heuristic} and Section~\ref{sec:ilp-model}---we describe a heuristic from the literature and phrase an ILP model for the tackled problem. Next, the proposed algorithm is outlined in Section~\ref{sec:algorithm}. Finally, an extensive experimental evaluation on artificial and real problem instances is provided in Section~\ref{sec:experiments}, and an outlook to future work is given in Section~\ref{sec:conclusions}.

\section{The LAPCS Problem}
\label{sec:lapcs}

Given two input sequences $x$ and $y$, we define the set of possible assignments $A$ as the set of all $a_{i,j}$---where $i \in \{1, \ldots, l_x\}$ and $j \in \{1, \ldots, l_y\}$---such that $x[i] = y[j]$. That is $A$ consists of all $a_{i,j}$ such that the letter at position $i$ of $x$ is equal to the letter at position $j$ of $y$. A valid common subsequence of the two input sequences $x$ and $y$ can then be represented by a subset $S \subseteq A$ that fullfills the following conditions:
\begin{itemize}
  \item {\bf Common subsequence condition:} For any two assignments $a_{i,j}, a_{k,l} \in S$ (where $a_{i,j} \not= a_{k,l}$) it must hold that either $i < k$ and $j < l$ or $i > k$ and $j > l$.
\end{itemize}
In order to translate such a solution into the corresponding common subsequence, the assignments in $S$ have to ordered from small to large indecees, either according to the first or the second index. Then, the letters corresponding to the assignments must be joined in this order. \\

A solution $S$ that fullfills the common subsequence condition is called \emph{arc-preserving} if the arcs induced by the solution are preserved: 
\begin{itemize}
  \item {\bf Arc preservation condition:} for any two assignments $a_{i,j}, a_{k,l} \in A$ (where $a_{i,j} \not= a_{k,l}$ and $i < k$) it must hold that $(i,k) \in P_x \Leftrightarrow (j,l) \in P_y$.
\end{itemize}

Given two arc-annotated input strings $x$ and $y$, the LAPCS problem consists in finding a solution $S \subseteq A$ that fullfills both the common subsequence and the arc preservation condition and is of maximal cardinality. Note that such a mapping corresponds to the longest arc-preserving common subsequence of $x$ and $y$. In~\cite{Evans1999,evans1999algorithms} it was shown that this problem is, in general, NP-hard. However, the structure of the arc annotation in the context of RNA sequences, for example, is in practise likely to satisfy some constraints. Concerning a string $x$, the following types of constraints have been considered in the literature:
\begin{enumerate}
  \item No two arcs may share an endpoint. That is, $\forall (i_1, i_2), (i_3, i_4) \in P_x$ it must hold that $i_1 \not= i_4$ and $i_2 \not= i_3$.
  \item Crossing arcs do not exist. That is, $\forall (i_1, i_2), (i_3, i_4) \in P_x$ it must hold that $i_3 \leq i_1 \leq i_4 \Leftrightarrow i_3 \leq i_1 \leq i_4$.
  \item Nesting arcs do not exist. That is, $\forall (i_1, i_2), (i_3, i_4) \in P_x$ it must hold that $i_1 \leq i_3 \Leftrightarrow i_2 \leq i_3$.
  \item No arcs exist. That is, $P_x = \emptyset$.
\end{enumerate}
Based on these four restrictions concerning the arc annotation, input strings are characterized into \textsc{unlimited} (no constraints), \textsc{crossing} (constraint 1), \textsc{nested} (constraints 1 and 2), \textsc{chain} (constraints 1, 2, and 3), and \textsc{plain} (constraint 4). Different versions of the LAPCS problem can therefore be denoted as follows: LAPCS($\cdot,\cdot$) where each of the two dots must be replaced by the characterization of the arc annotation of the first and the second input string. For example, in problem LAPCS(\textsc{unlimited},\textsc{nested}), the arc annotation of the first input string is characterized as \textsc{unlimited}, and the one of the second one as \textsc{nested}. It is well known that LAPCS(\textsc{plain},\textsc{plain}), for example, can be solved in polynomial time with the dynamic programming algorithm by Smith and Waterman~\cite{smith1981identification}. In this paper, however, we deal with the most general version of the problem, LAPCS(\textsc{unlimited},\textsc{unlimited}), which---as mentioned above---was shown to be NP-hard. For simplicity reasons we refer to this problem version simply as LAPCS. \\

\section{Existing Heuristic for LAPCS}
\label{sec:heuristic}

As far as we know, the only heuristic from the literature that is applicable to the most general version of the LAPCS problem was described in~\cite{Jiang2004257}, and works as follows. First, the dynamic programming algorithm by Smith and Waterman is applied to input strings $x$ and $y$, disregarding the arc annotations. The result is a mapping $S \subseteq A$ that---most probably---violates some of the arc preservation constraints. In order to \emph{repair} this invalid solution, the following is done. First a graph $G$ is constructed as follows. A vertex $v$ is introduced for each assignment $a_{i,j} \in S$. Two vertices $v$ (corresponding to an assignment $a_{i,j} \in S$) and $v'$ (corresponding to an assignment $a_{k,l} \in S$ with $i < k$) are connected by an edge if either $(i, k) \in P_x$ or $(j, l) \in P_y$, but not both. In other words, two vertices are connected by an edge if they represent a violation of the arc preservation constraints. Note that in order to repair $S$ by removing as few assignments from $S$ as possible, we can solve the \emph{maximum independent set} (MIS) problem in $G$, and remove all assignments from $S$ that correspond to vertices that are not in the optimal solution to the MIS problem. In our implmentation we used CPLEX 12.6 to solve the MIS problem in all cases.

\section{An ILP Model for LAPCS}
\label{sec:ilp-model}

The LAPCS problem can be stated in terms of an integer linear program (ILP) in the following way. For each $a_{i,j} \in A$ is introdcued a binary variable $z_{i,j}$. The set of all binary variables is denoted by $Z$. We say that two variables $z_{i,j} \not= z_{k,l}$ (where $i \leq k$) are \emph{in conflict}, if setting both variables to one violates (1) the common subsequence condition, (2) the are preservation condition, or both. In technical terms, two variables $z_{i,j} \not= z_{k,l}$ (where $i \leq k$) are in conflict, if at least one of the following holds:
\begin{enumerate}
  \item $j \geq l$
  \item Either $(i,k) \in P_x$ or $(j,l) \in P_y$, but not both at the same time.
\end{enumerate}
The LAPCS problem can then be rephrased as the problem of selecting a maximal number of non-conflicting variables from $Z$. Given these notations, the ILP is stated as follows.

\begin{empheq}[box=\shadowbox*]{align}
  & \mbox{\bf max} \sum_{z_{i,j} \in Z} z_{i,j} \\
  \shoveright{\text{\bf subj.~to:}} & \nonumber \\
  z_{i,j} + z_{k,l}                   & \leq 1   \;\;\; \forall \; z_{i,j} \not= z_{k,l}, i \leq k \mbox{ in conflict} \label{eqn:const2} \\
  z_{i,j}                             & \in \{0,1\} \;\;\; \mbox{for } z_{i,j} \in Z
\end{empheq}

Hereby, constraints (\ref{eqn:const2}) ensure that selected variables are not in conflict.

\section{The Hybrid EA with Solution Merging}
\label{sec:algorithm}

The proposed hybrid EA, henceforth labelled H\textsc{yb}-E\textsc{a}, is pseudo-coded in Algorithm~\ref{algo:ea}. In the context of this algorithm, valid solutions to the problem are subsets of the complete set $Z$ of variables introduced in the context of the ILP model. If a solution $S$ contains a variable $z_{i,j}$, this means that the variable must be given value one in order to produce the corresponding solution. The main loop of the EA is executed while the CPU time limit is not reached. It consists of the following actions. First, the best-so-far solution $S_{\mbox{\tiny bsf}}$ is initialized to $\emptyset$. Then, at each iteration, first, the set of variables representing a set of merged solutions is initialized with the best-so-far solution $S_{\mbox{\tiny bsf}}$. Then, a number of $n_{\mathrm{sols}}$ solutions is probabilistically constructed in function \textsf{GenerateRandomSolution}($\drate$, $\lsize$, $Z$) in line 6 of Algorithm~\ref{algo:ea}. The variables contained in these solutions are added to $S'$. Afterwards, solution merging is applied to $S'$, that is, an ILP solver is applied to find the best valid solution that can be built from the variables in $S'$ (see function \textsf{ApplySolutionMerging}($t_{\max}$, $S'$) in line~9 of Algorithm~\ref{algo:ea}). Parameter $t_{\max}$ is a time limit for the ILP solver. In particular, the output of this function is the best solution found by the ILP solver within $t_{\max}$ seconds. Note that for applying an ILP solver to $S' \subseteq Z$, all the appearences of $Z$ in the ILP model of Section~\ref{sec:ilp-model} have to be replaced with $S'$. In case $S^{\prime}_{\mathrm{opt}}$ is better than the current best-so-far solution $S_{\mathrm{bsf}}$, solution $S^{\prime}_{\mathrm{opt}}$ is stored as the new best-so-far solution (line~10). The output of the algorithm is the best-so-far solution $S_{\mbox{\tiny bsf}}$. \\

\begin{algorithm}[h]
\caption{H\textsc{yb}-E\textsc{a} for the LAPCS problem} \label{algo:ea}
\begin{algorithmic}[1]
\STATE {\bf input:} strings $x$ and $y$ over alphabet $\Sigma$, values for parameters $n_{\mathrm{sols}}$, $\drate$, $\lsize$, and $t_{\max}$
\STATE $S_{\mathrm{bsf}} := \emptyset$
\WHILE{CPU time limit not reached}
    \FOR{$i = 1, \ldots, n_{\mathrm{sols}}$}
        \STATE $S' := S_{\mathrm{bsf}}$
        \STATE $S :=$ \textsf{GenerateRandomSolution}($\drate$, $\lsize$, $Z$)
        \STATE $S' := S' \cup S$
    \ENDFOR
    \STATE $S^{\prime}_{\mathrm{opt}} :=$ \textsf{ApplySolutionMerging}($t_{\max}$, $S'$)
    \STATE {\bf if} $|S^{\prime}_{\mathrm{opt}}| > |S_{\mathrm{bsf}}|$ {\bf then} $S_{\mathrm{bsf}} := S^{\prime}_{\mathrm{opt}}$
\ENDWHILE
\STATE {\bf output:} $S_{\mathrm{bsf}}$
\end{algorithmic}
\end{algorithm}

In the following we will describe in detail the remaining component of the algorithm: the probabilistic construction of solutions in function \textsf{GenerateRandomSolution}($\drate$, $\lsize$, $Z$). First, a common subsequence of $x$ and $y$ is---without regarding the arc preservation constraints---probabilistically generated as follows. For this purpose let us first introduce for each letter $a \in \Sigma$ the subset $Z_a \subseteq Z$ of variables which correspond to letter $a$. A solution construction starts with an empty solution $S = \emptyset$, and the first step consists in generating the set of variables $C \subseteq Z$ that serve as options to be added to $S$. More specifically, the initial set $C$ is generated in order to contain for each letter $a \in \Sigma$ the variable $z_{i,j} \in Z_a$ (if any) such that $i \leq k$ and $j \leq l$, $\forall z_{k,l} \in Z_a$. Moreover, options $z_{i,j} \in C$ are given a weight value $w(z_{i,j}) := \frac{i}{l_x} + \frac{j}{l_y}$, which is a known greedy function for longest common subsequence problems (see, for example,~\cite{Fra95:phd,HuaYanTse04:ics}). At each construction step, exactly one variable is chosen from $C$ and added to $S$. For doing so, first, a value $r$ is chosen uniformly at random from $[0,1]$. In case $r \leq \drate$, where $\drate$ is a parameter of the algorithm, the variable $z_{i,j} \in C$ with the smallest weight value is deterministically chosen. Otherwise, a candidate list $L \subseteq C$ of size $\min\{\lsize, |C|\}$ containing the options with the lowest weight values is generated and exactly one variable $z_{i,j} \in L$ is then chosen uniformly at random and added to $S$. Note that $\lsize$ is another parameter of the solution construction process. Finally, the set of options $C$ for the next construction step is generated. Being $z_{i,j}$ the last variable that was added to $S$, $C$ contains for each letter $a \in \Sigma$ the variable $z_{r,s} \in Z_a$ (if any) with the lowest weight value $w(z_{r,s})$ calculated as $w(z_{r,s}) := \frac{r-i}{l_x-i} + \frac{s-j}{l_y-j}$. The solution construction is finished when the set of options is empty. 

However, note that a solution $S$ constructed in the way as described above does not necesarrily respect all arc preservation constraints. Therefore, the same \emph{repair mechanism} is utilized as in the heuristic from Section~\ref{sec:heuristic} in order to transform $S$ into a valid LAPCS solution.

\section{Experimental Evaluation}
\label{sec:experiments}

Summarizing, the following techniques are included in the experimental evaluation: (1) the heuristic described in Section~\ref{sec:heuristic} (H\textsc{euristic}), (2) the hybrid EA (\textsc{Hyb-Ea}), and (3) \textsc{Hyb-Ea} without the application of solution merging. This last algorithm---henceforth denoted by \textsc{Ms-Heur}---is bascially a multi-start heuristic that constructs randomized solutions (and applies the repair procedure to them) until it runs out of computation time. Comparing \textsc{Hyb-Ea} with \textsc{Ms-Heur} will enable us to measure the contribution of solution merging. The three above-mentioned algorithms were implemented in ANSI C++ using GCC 4.7.3, without the use of any external libraries. In addition, the ILP models in the context of \textsc{Hyb-Ea} were solved with the ILP solver IBM ILOG CPLEX v12.6 in one-threaded mode. The experimental evaluation has been performed on a cluster of PCs with Intel(R) Xeon(R) CPU 5670 CPUs of 12 nuclei of 2933 MHz and at least 40 Gigabytes of RAM. Note that we also tried to apply CPLEX to the complete ILP models for each problem instance. However, the models were too large, even in the case of the smallest problem instances.

The remainder of this section is organized as follows. First, the set of benchmark instances is described. Second, the tuning experiments that were conducted in order to determine a proper setting for the parameters of \textsc{Hyb-Ea} are outlined. Finally, an exhaustive experimental evaluation is presented.

\subsection{Benchmark Instances}
\label{sec:instances}

Two sets of benchmark instances were generated. The first set, labelled \textsc{Set1}, consists of artificial problem instances. Each of these instances consists of two artifically generated RNA strings of length $n \in \{100, 200, \ldots, 900, 1000\}$. The probability of each letter and each position was chosen to be $1/4$. Moreover, for each input string we randomly generated a number of $n_{\mathrm{arcs}} \in \{n/10, n/5, n/2\}$. Hereby, it was taken care that all $n_{\mathrm{arcs}}$ arcs were different. For each combination of $n$ and $n_{\mathrm{arcs}}$ we randomly generated 30 problem instances. This makes a total of 900 problem instances.

\begin{figure*}[!h]
\centering
\subfloat[\label{fig:real:a}]{
  \includegraphics[width=0.46\textwidth]{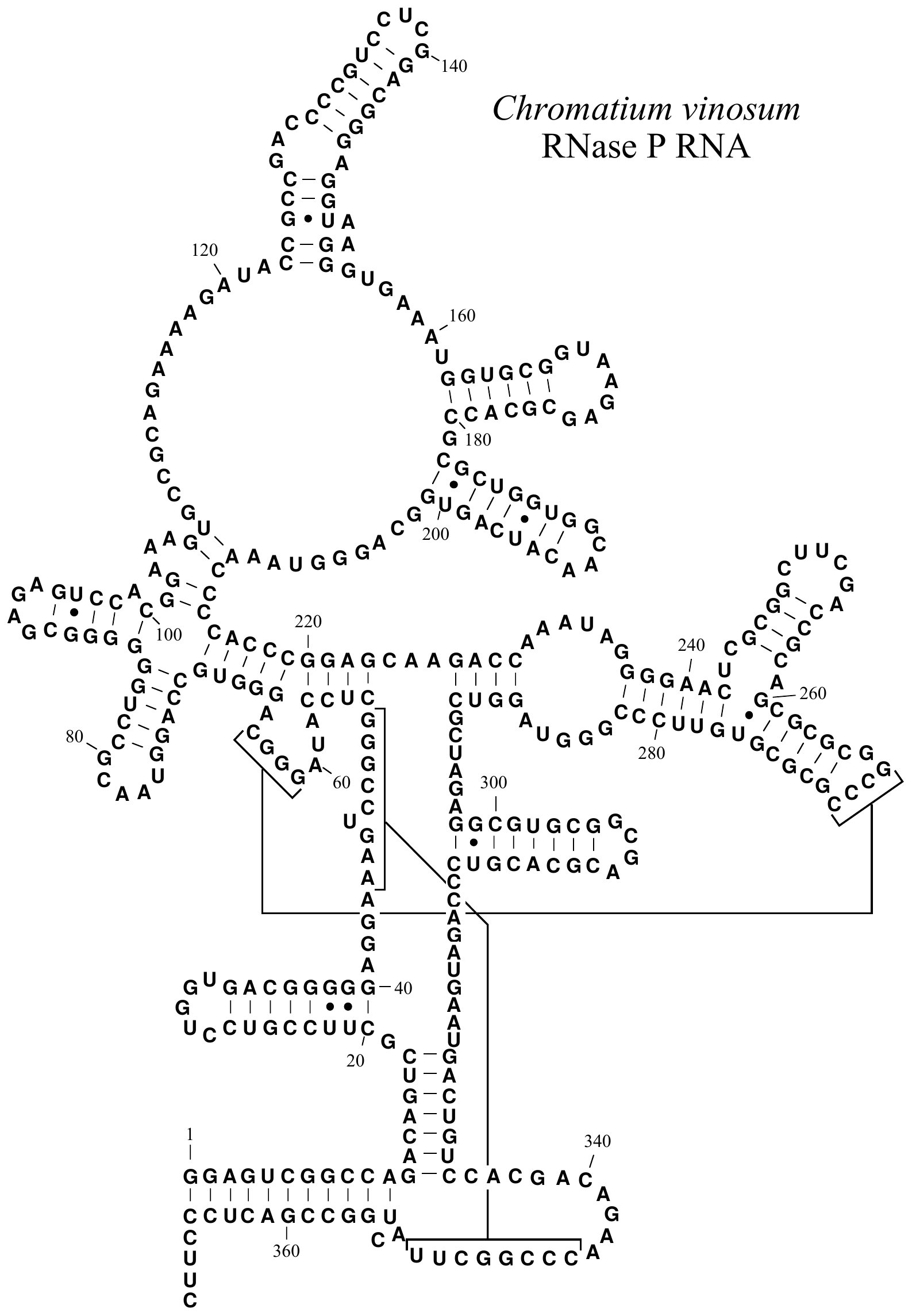}
} 
\subfloat[\label{fig:real:b}]{
  \includegraphics[width=0.46\textwidth]{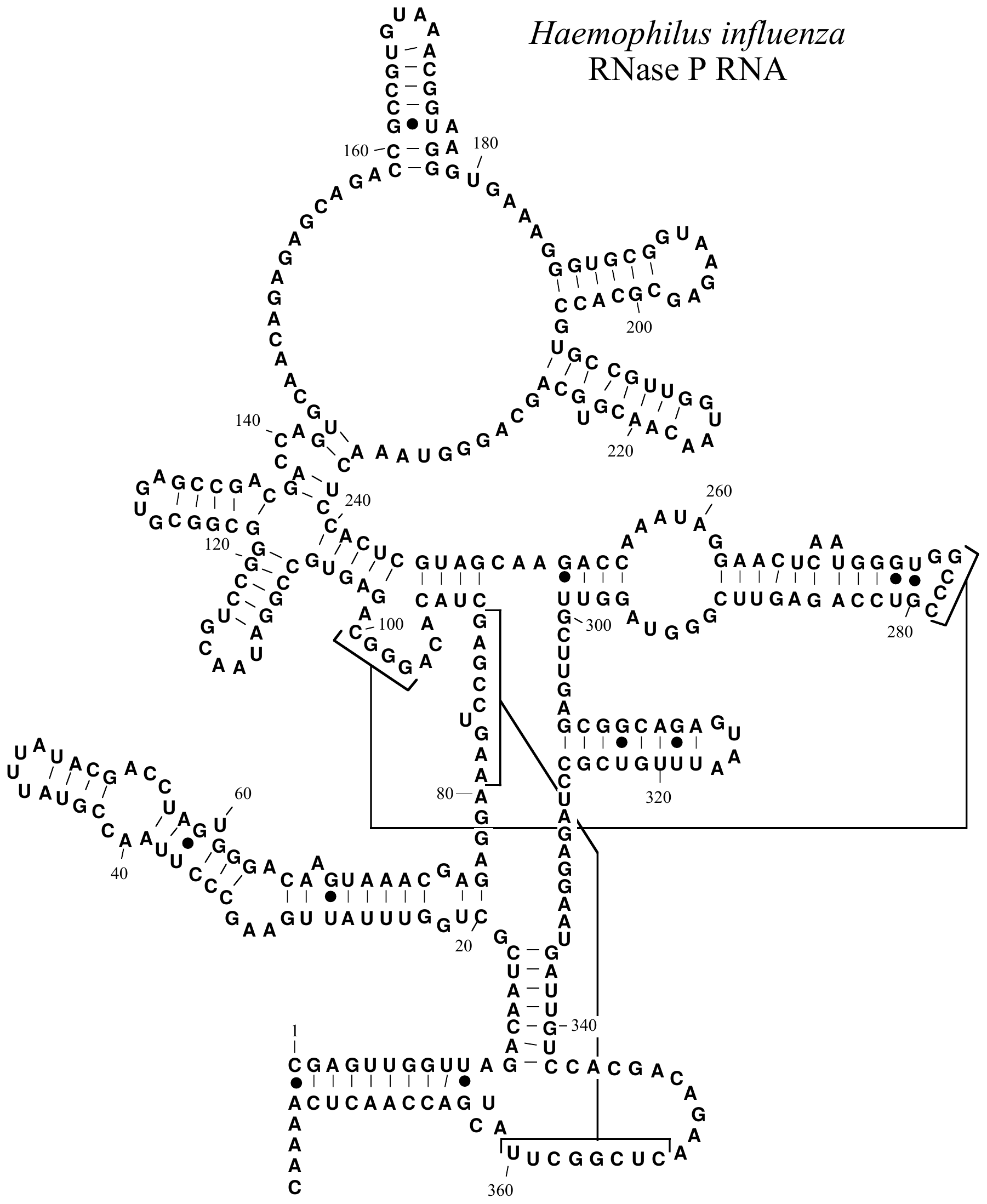}
} \\
\subfloat[\label{fig:real2:a}]{
  \includegraphics[width=0.46\textwidth]{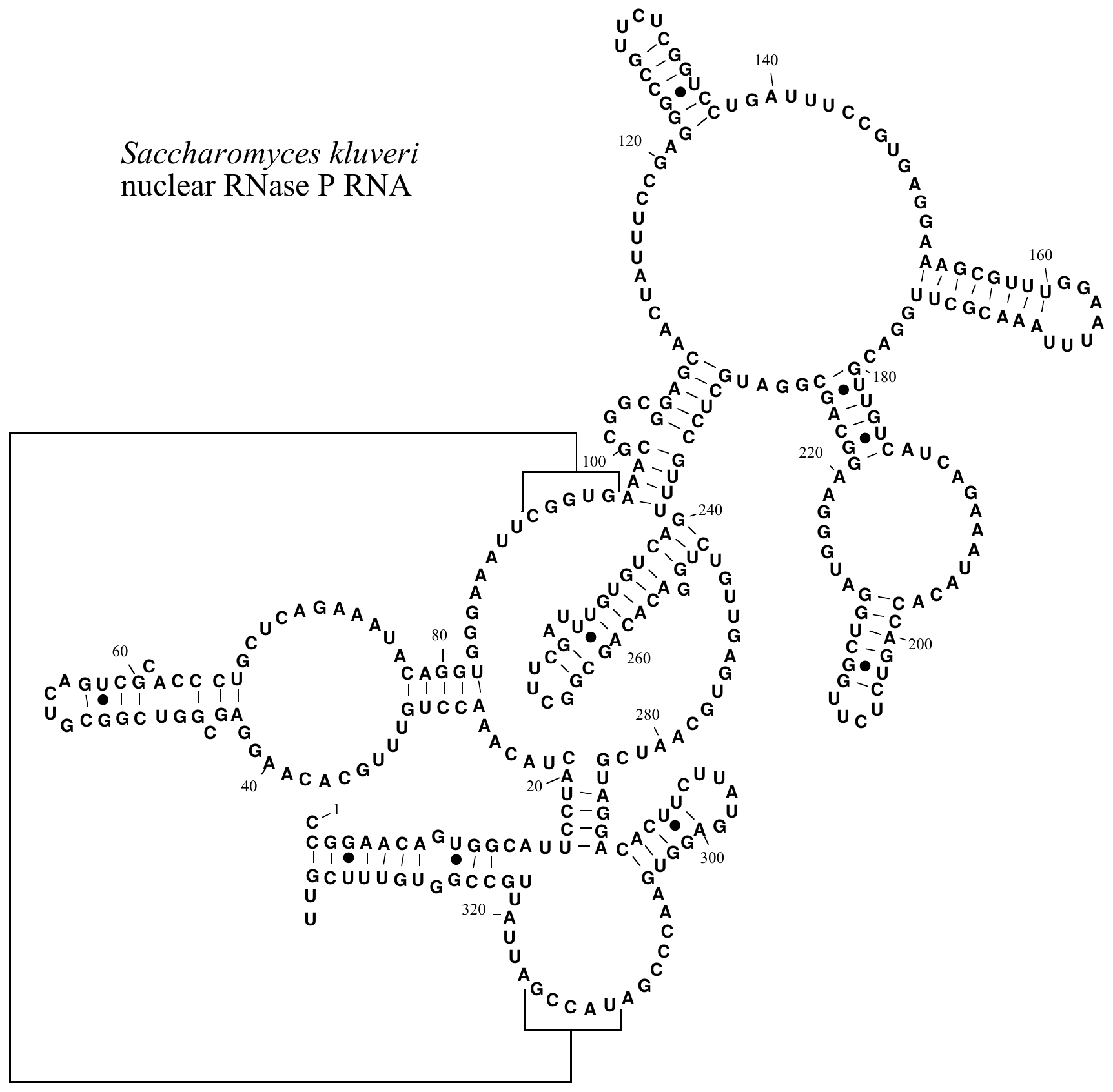}
} 
\subfloat[\label{fig:real2:b}]{
  \includegraphics[width=0.46\textwidth]{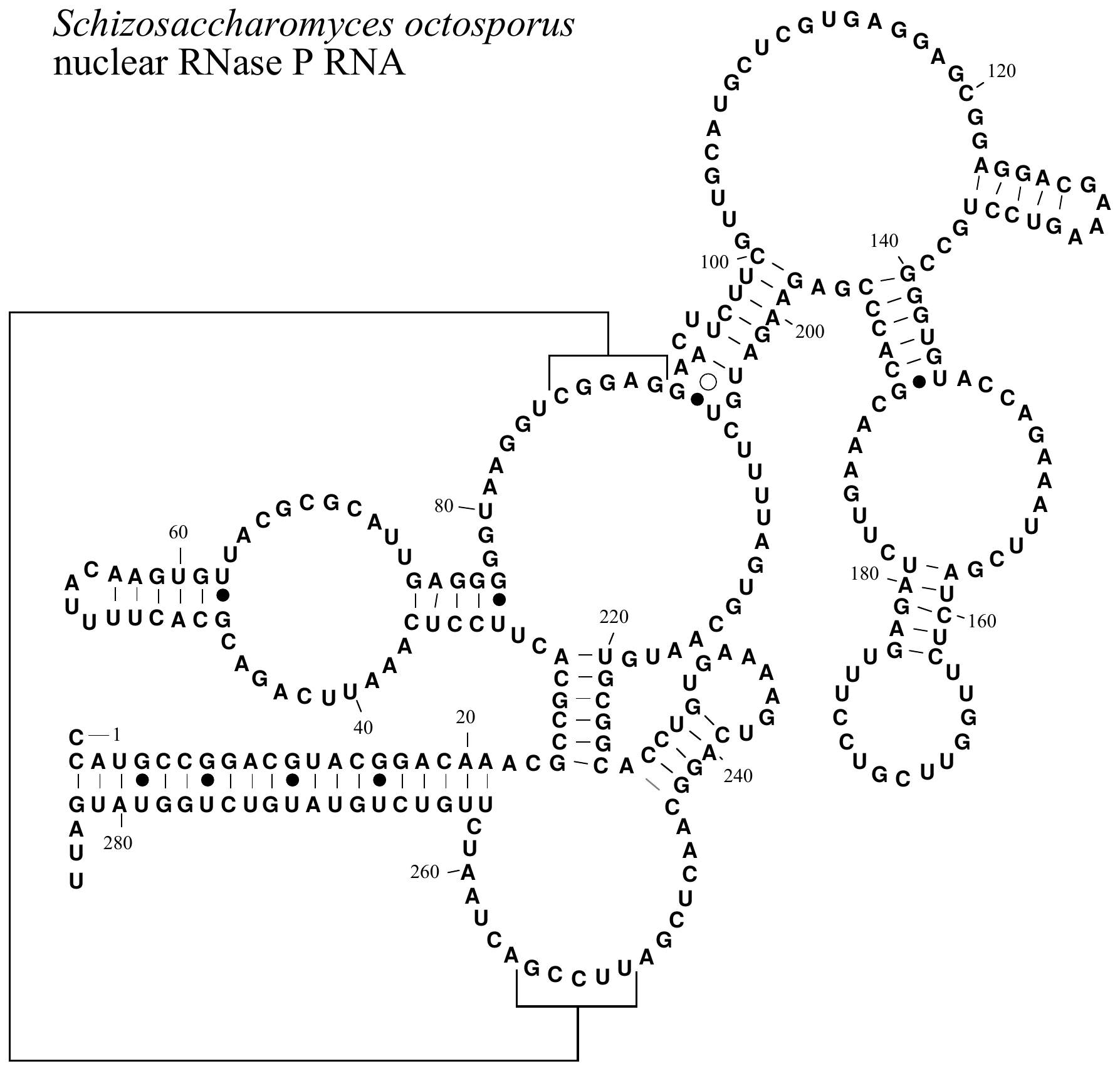}
}
\caption{Secondary structure of the two RNA sequences involved in instances Real\_1 (a and b) and Real\_8 (c and d). All graphics were downloaded from the RNase P Database~\cite{brown1999ribonuclease} (a) RNA of \emph{Allochromatium vinosum}, (b) RNA of \emph{Haemophilus influenza}, (c) RNA of \emph{Saccharomyces kluveri}, (d) RNA of \emph{Schizosaccharomyces octosporus}.}
\label{fig:real}
\end{figure*}

For the second benchmark set, labelled \textsc{Set2}, we downloaded arc-annotated RNA sequences from the RNase P Database~\cite{brown1999ribonuclease}. In total we assembled 10 problem instances, whose characteristics are described in Table~\ref{tab:instance-characteristics}. Moreover, the secondary structures of the RNA sequences involved in instances Real\_1 and Real\_8 are exemplary shown in Figure~\ref{fig:real}.

\begin{table*}[!t]
\caption{Characteristics of the real-life instances. All 20 RNA sequences, together with their secondary structure, were downloaded from the RNase P Database~\cite{brown1999ribonuclease}.}
\label{tab:instance-characteristics}
\centering
\scalebox{0.75}{
\begin{tabular}{lllllllll} 
\hline
{\bf Instance} &$\;$& \multicolumn{3}{c}{\bf First String}  &$\;$& \multicolumn{3}{c}{\bf Second string} \\ 
{\bf name}     &    & {\bf RNA} & {\bf Lenght} & {\bf Arcs} &    & {\bf RNA} & {\bf Lenght} & {\bf Arcs} \\ \cline{1-1} \cline{3-5} \cline{7-9}
Real\_1        &    & \emph{Allochromatium vinosum}       & 369 & 119    &    & \emph{Haemophilus influenza}          & 377  & 124  \\
Real\_2        &    & \emph{Bacteroides thetaiotaomicron} & 361 & 121    &    & \emph{Porphyromonas gingivalis}       & 398  & 131  \\
Real\_3        &    & \emph{Halococcus morrhuae}          & 475 & 154    &    & \emph{Haloferax volcanii}             & 433  & 142  \\
Real\_4        &    & \emph{Klebsiella pneumoniae}        & 383 & 127    &    & \emph{Escherichia coli}               & 377  & 124  \\
Real\_5        &    & \emph{Methanococcus jannaschii}     & 252 & 75     &    & \emph{Archaeoglobus fulgidus}         & 229  & 67   \\
Real\_6        &    & \emph{Methanosarcina barkeri}       & 371 & 115    &    & \emph{Pyrococcus abyssi}              & 330  & 100  \\
Real\_7        &    & \emph{Mycoplasma genitalium}        & 384 & 119    &    & \emph{Mycoplasma pneumoniae}          & 369  & 112  \\
Real\_8        &    & \emph{Saccharomyces kluveri}        & 336 & 90     &    & \emph{Schizosaccharomyces octosporus} & 281  & 71   \\
Real\_9        &    & \emph{Serratia marcescens}          & 378 & 125    &    & \emph{Shewanella putrefaciens}        & 354  & 115  \\
Real\_10       &    & \emph{Streptomyces bikiniensis}     & 398 & 135    &    & \emph{Streptomyces lividans}          & 405  & 138  \\ \hline
\end{tabular}}
\end{table*}

\subsection{Algorithm Tuning}
\label{sec:tuning}

The automatic configuration tool \textsf{irace}~\cite{LopezIbanez201643} was used for tuning the parameters of \textsc{Hyb-Ea}. The following parameters of \textsc{Hyb-Ea} were considered for tuning: ($n_{\mathrm{sols}}$) the number  of solution constructions per iteration ($\drate$) the determinism rate, ($\lsize$) the candidate list size, and ($t_{\max}$) the maximum time in seconds allowed for solution merging (at each call of the solution merging procedure). In particular, \textsc{Hyb-Ea} was tuned separately for each input string length, which---after initial experiments---seemed to have a greater influence on the behavior of the algorithm than the number of arcs. For each $n \in \{100, 200, \ldots, 900, 1000\}$ we randomly generated two tuning instances for each of the three values of $n_{\mathrm{arcs}}$. This makes a total of six tuning instances for each $n$. The tuning process for each $n$ was given a budget of 1000 runs of \textsc{Hyb-Ea}, where each run was given a computation time limit of $n/10$ CPU seconds. Finally, the following parameter value ranges were considered concerning the four parameters of \textsc{Hyb-Ea}:
\begin{itemize}
  \item $n_{\mathrm{sols}} \in \{5, 10, 20\}$
  \item $\drate \in \{0.0, 0.3, 0.5, 0.7, 0.9\}$, where a value of $0.0$ means that the selection of the assignment to be added to the partial solution under construction is always done randomly from the candidate list, while a value of $0.9$ means that solution constructions are nearly deterministic.
  \item $\lsize \in \{1, 2, 3, 4\}$
  \item $t_{\max} \in \{1.0, 5.0, 10.0, 20.0\}$ (in seconds).
\end{itemize}
The tuning runs with \textsf{irace} produced the configurations of \textsc{Hyb-Ea} as shown in Table~\ref{tab:parameter-values:hyb-ea}. The following trends can be observed. Apart from $n=100$, the number of solution constructions is always set to five. This is because the smaller $n_{\mathrm{sols}}$, the smaller is the ILP model that has to be solved by CPLEX in the context of solution merging at each iteration of the algorithm. Moreover, the smaller the ILP model, the more efficient is CPLEX in solving such a model. The values of $\drate$ are consistently between $0.3$ and $0.7$, whereas the values of $\lsize$ are consistently set to two or three. Finally, the settings of $t_{\max}$ seem somewhat erratic. However, this is due to the fact that the application of CPLEX in solution merging is very efficient and stops, most of times, much below 5 CPU seconds. Therefore, it is only of importance that the value of $t_{\max}$ is at least set to 5.0.

\begin{table}[!t]
\caption{Results of tuning \textsc{Hyb-Ea} with \textsf{irace}.}
\label{tab:parameter-values:hyb-ea}
\centering
\begin{tabular}{>{\centering}m{1cm}|rrrr} \hline
$n$       & $n_{\mathrm{sols}}$ & $\drate$ & $\lsize$ & $t_{\max}$ \\ \hline
100       & 10                & 0.3        & 2        & 5.0       \\
200       & 5                 & 0.7        & 3        & 1.0       \\
300       & 5                 & 0.7        & 2        & 5.0       \\
400       & 5                 & 0.7        & 3        & 10.0      \\
500       & 5                 & 0.3        & 2        & 20.0      \\
600       & 5                 & 0.7        & 2        & 5.0       \\
700       & 5                 & 0.5        & 2        & 20.0      \\
800       & 5                 & 0.7        & 2        & 5.0       \\
900       & 5                 & 0.5        & 2        & 5.0       \\
1000      & 5                 & 0.7        & 2        & 5.0       \\ \hline
\end{tabular}
\end{table}

\subsection{Numercial Results}
\label{sec:results}

\begin{table*}[!t]
\caption{Experimental results concerning the artificial problem instances from \textsc{Set1}.}
\label{tab:results:artificial}
\centering
\scalebox{1.0}{
\begin{tabular}{ccrrrrrrrr} 
\hline
$n$ & \# arcs & \multicolumn{2}{c}{\textsc{Heuristic}} &$\;$& \multicolumn{2}{c}{\textsc{Ms-Heur}} &$\;$& \multicolumn{2}{c}{\textsc{Hyb-Ea}} \\ \cline{3-4} \cline{6-7} \cline{9-10}
    &        & \textbf{result} & \textbf{time} && \textbf{result} & \textbf{time} && \textbf{result} & \textbf{time} \\ \hline
\multirow{3}{*}{$100$}   & 10  &  55.73   & $<1$ &&  51.80   & 3.70  && \ccg58.87   & 1.17 \\
                         & 20  &  51.63   & $<1$ &&  49.23   & 4.10  && \ccg57.00   & 1.52 \\
                         & 50  &  42.63   & $<1$ &&  42.57   & 3.84  && \ccg50.07   & 2.26 \\ \hline
\multirow{3}{*}{$200$}   & 20  &  113.77  & $<1$ &&  104.33  & 7.74  && \ccg120.57  & 7.36 \\
                         & 40  &  104.13  & $<1$ &&  97.67   & 11.06 && \ccg114.97  & 7.22 \\
                         & 100 &  87.10   & $<1$ &&  84.20   & 6.19  && \ccg101.70  & 7.66 \\ \hline
\multirow{3}{*}{$300$}   & 30  &  170.43  & $<1$ &&  158.70  & 12.87 && \ccg178.77  & 13.18 \\
                         & 60  &  156.87  & $<1$ &&  149.80  & 11.55 && \ccg171.47  & 12.08 \\
                         & 150 &  132.40  & $<1$ &&  128.17  & 15.45 && \ccg151.67  & 13.82 \\ \hline
\multirow{3}{*}{$400$}   & 40  &  229.20  & $<1$ &&  198.27  & 17.05 && \ccg239.53  & 24.43 \\
                         & 80  &  210.93  & $<1$ &&  185.23  & 17.37 && \ccg228.97  & 27.74 \\
                         & 200 &  175.63  & $<1$ &&  159.77  & 18.10 && \ccg202.60  & 28.26 \\ \hline
\multirow{3}{*}{$500$}   & 50  &  289.20  & $<1$ &&  235.90  & 19.25 && \ccg298.43  & 32.93 \\
                         & 100 &  263.80  & $<1$ &&  222.83  & 22.02 && \ccg285.83  & 33.89 \\
                         & 250 &  220.43  & $<1$ &&  192.57  & 21.52 && \ccg253.13  & 39.81 \\ \hline
\multirow{3}{*}{$600$}   & 60  &  346.63  & $<1$ &&  310.50  & 26.86 && \ccg360.10  & 32.62 \\
                         & 120 &  317.93  & $<1$ &&  290.03  & 26.07 && \ccg343.70  & 44.54 \\
                         & 300 &  266.03  & $<1$ &&  247.00  & 25.24 && \ccg304.43  & 43.58 \\ \hline
\multirow{3}{*}{$700$}   & 70  &  404.53  & $<1$ &&  343.33  & 25.42 && \ccg420.77  & 46.32 \\
                         & 140 &  369.97  & $<1$ &&  320.97  & 28.49 && \ccg398.37  & 51.93 \\
                         & 350 &  308.57  & $<1$ &&  275.17  & 31.20 && \ccg352.50  & 57.58 \\ \hline
\multirow{3}{*}{$800$}   & 80  &  461.17  & $<1$ &&  408.50  & 34.41 && \ccg478.97  & 59.30 \\
                         & 160 &  424.57  & $<1$ &&  380.97  & 35.10 && \ccg458.10  & 61.30 \\
                         & 400 &  353.57  & $<1$ &&  326.33  & 36.55 && \ccg403.73  & 65.53 \\ \hline
\multirow{3}{*}{$900$}   & 90  &  520.87  & $<1$ &&  435.33  & 39.06 && \ccg537.53  & 72.56 \\
                         & 180 &  477.90  & $<1$ &&  409.20  & 44.19 && \ccg513.90  & 71.06 \\
                         & 450 &  398.97  & $<1$ &&  350.83  & 48.57 && \ccg453.20  & 75.98 \\ \hline
\multirow{3}{*}{$1000$}  & 100 &  578.30  & $<1$ &&  507.60  & 42.37 && \ccg599.63  & 76.23 \\
                         & 200 &  531.33  & $<1$ &&  473.40  & 43.07 && \ccg570.93  & 78.28 \\
                         & 500 &  443.33  & $<1$ &&  404.23  & 35.12 && \ccg505.37  & 84.39 \\ \hline
\end{tabular}}
\end{table*}

\begin{figure*}[!t]
\centering
\subfloat[\label{fig:improvement:a}]{
  \includegraphics[width=0.46\textwidth]{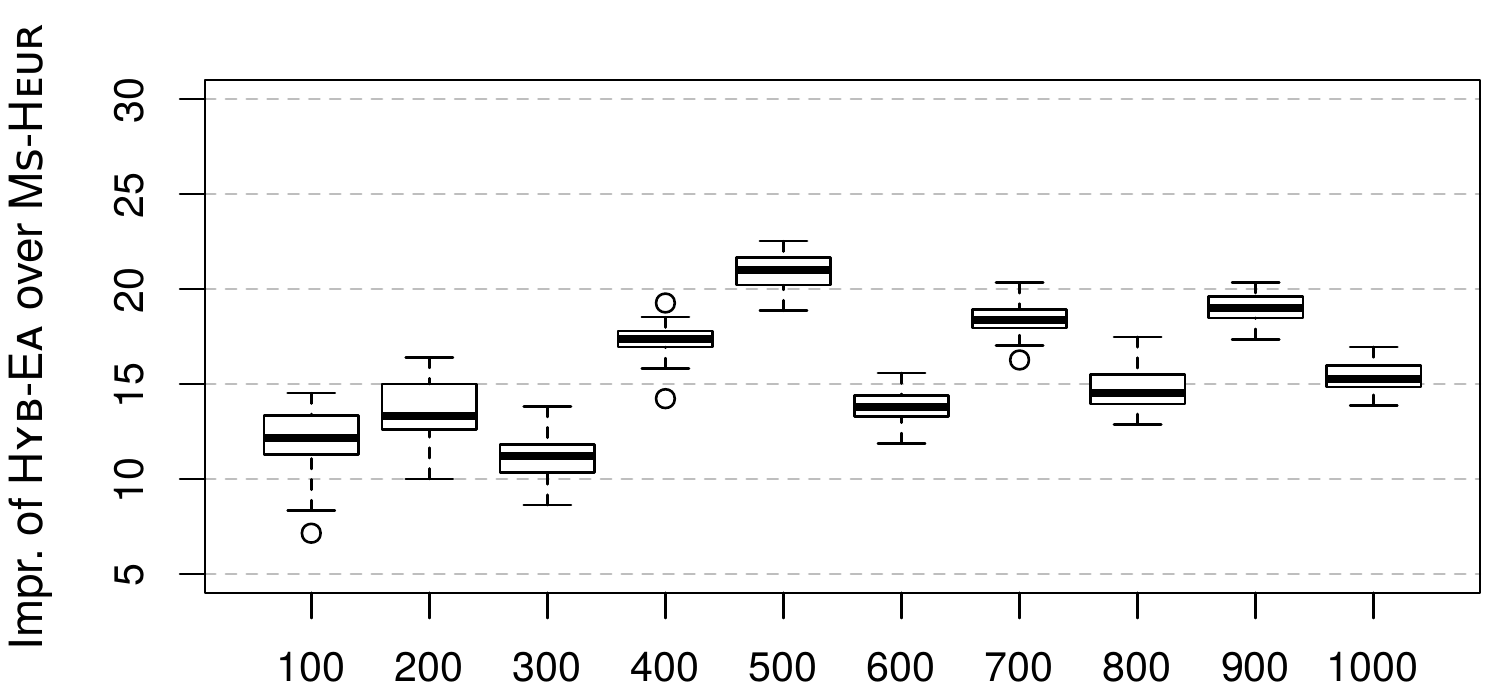}
} 
\subfloat[\label{fig:improvement:b}]{
  \includegraphics[width=0.46\textwidth]{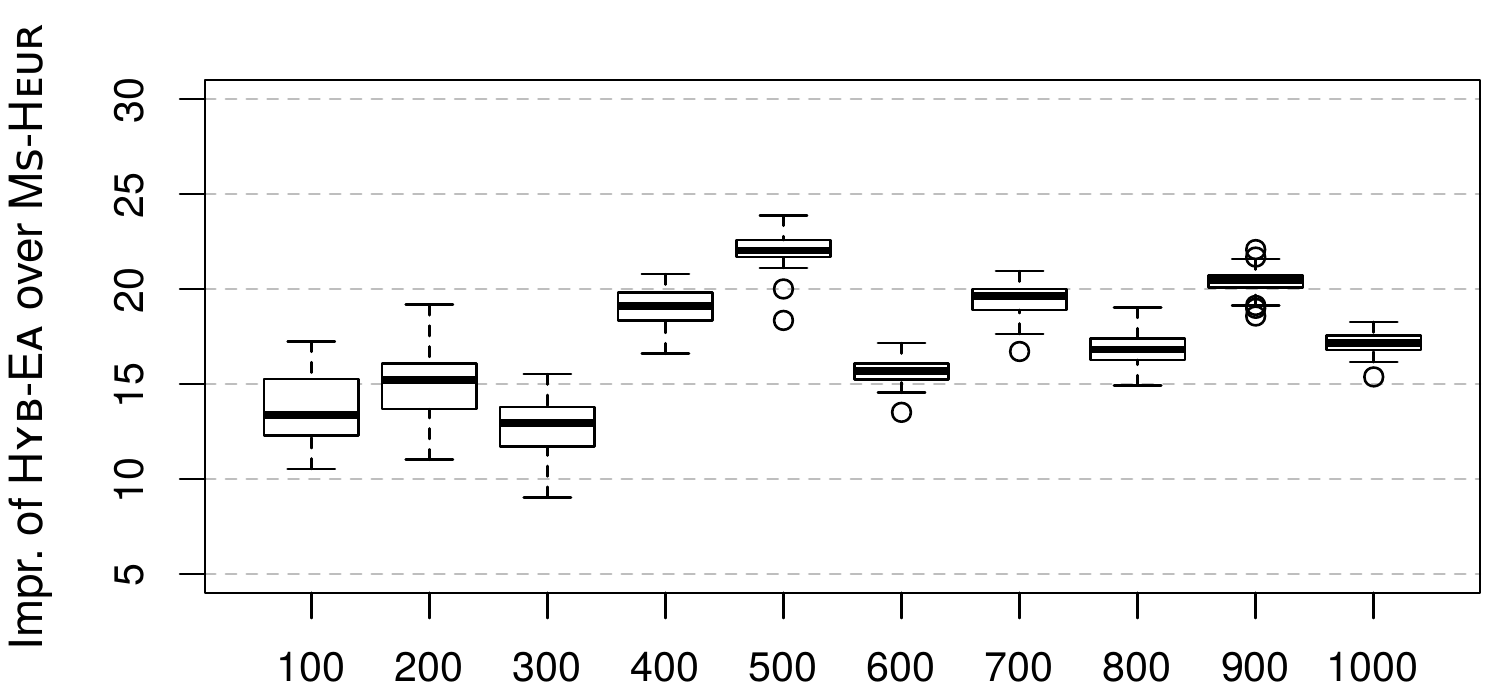}
} \\
\subfloat[\label{fig:improvement:c}]{
  \includegraphics[width=0.46\textwidth]{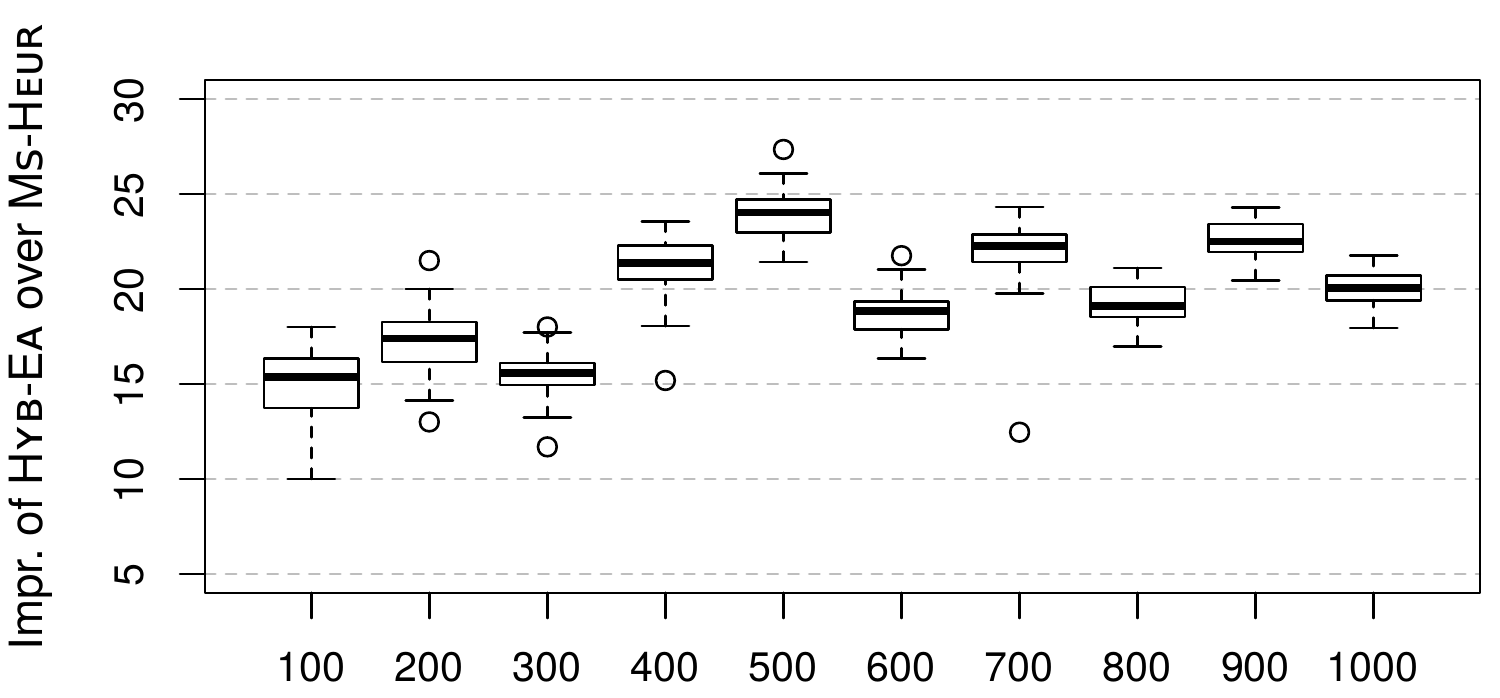}
} 
\caption{Improvement of \textsc{Hyb-Ea} over \textsc{Ms-Heur} (in percent). Each box shows the differences for the corresponding 30 instances. (a) Instances with $n_{\mathrm{arcs}} = n/10$, (b) instances with $n_{\mathrm{arcs}} = n/5$, (c) instances with $n_{\mathrm{arcs}} = n/2$.}
\label{fig:improvement}
\end{figure*}

The results concerning the artificial problem instances from \textsc{Set1} are presented in Table~\ref{tab:results:artificial}. Each row provides the results of \textsc{Heuristic}, \textsc{Ms-Heur} and \textsc{Hyb-Ea} in terms of the average solution quality obtained for the 30 problem instances of the corresponding combination of $n$ and $n_{\mathrm{arcs}}$. All techniques were applied exactly once to each problem instance. The computation time limit used for \textsc{Ms-Heur} and \textsc{Hyb-Ea} was $n/10$ CPU seconds. The column with heading \textbf{time} shows the average computation time for the 30 problem instances in the case of \textsc{Heuristic}, and the average time at which the best solution of a run was found, in the case of \textsc{Ms-Heur} and \textsc{Hyb-Ea}. The best result of each table row is marked by a lightgrey background. The following observations can be made:
\begin{itemize}
  \item \textsc{Heuristic} is very fast. Its application to any of the problem instances requires less than one CPU second. Moreover, the results of \textsc{Heuristic} are always better than the results of \textsc{Ms-Heur}. 
  \item \textsc{Hyb-Ea} is, by far, the best algorithm in the comparison. It obtains the best result for each combination of $n$ and $n_{\mathrm{arcs}}$. In particular, \textsc{Hyb-Ea} always improves over \textsc{Ms-Heur}. This means that the solution merging component is an essential part of \textsc{Hyb-Ea}. This is remarkable as the application of CPLEX to the original problem instances was not viable at all. Nevertheless, in the context of solution merging, mathematical programming plays an important role for the success of the \textsc{Hyb-Ea} algorithm. 
\end{itemize}
In order to study the magnitude of the improvement of \textsc{Hyb-Ea} over \textsc{Ms-Heur}, we show the percentage improvement of \textsc{Hyb-Ea} over \textsc{Ms-Heur}---by means of boxplots---for each combination of $n$ and $n_{\mathrm{arcs}}$ in the three graphics of Figure~\ref{fig:improvement}. The x-axis of each graphic ranges from $n=100$ to $n=1000$. These grapics show, first, that the improvement of \textsc{Hyb-Ea} over \textsc{Ms-Heur} is generally around $15-20\%$. Moreover, it can be observed that the improvements become bigger with a growing number of arcs. \\

In a second set of experiments we applied the three algorithms considered in this work to the set of 10 real problem instances (\textsc{Set2}). The results are provided in Table~\ref{tab:results:real}, in the following way. \textsc{Heuristic} was applied exactly once to each problem instance. The corresponding result can be found in columns with headings {\bf result} and {\bf time}. Both \textsc{Ms-Heur} and \textsc{Hyb-Ea} are applied 30 times to each problem instance, with a computation time limit of 30 seconds per application. In both cases we provide the best result obtained over 30 applications (column {\bf best}), the average of the 30 results obtained (column {\bf avg.}), and the average time at which the best solution of each run was found (column {\bf time}). Again, the best result for each of the 10 instances is marked by a lightgrey background. Apart from one case (Real\_4), the results are consistent with our observations in the context of the artificial benchmark instances. In the case of Real\_4, \textsf{Heuristic} outperforms \textsc{Hyb-Ea}. This suggest that this problem instance has certain characteristics that are difficult to be discovered by the step-by-step way of constructing solutions as used in \textsc{Hyb-Ea}. Remember that \textsc{Heuristic}, in contrast, uses the dynamic programming algorithm by Smith and Waterman in order to generate the (possibly invalid) initial solution. 

\begin{table*}[!t]
\caption{Experimental results concerning the real problem instances from \textsc{Set2}.}
\label{tab:results:real}
\centering
\scalebox{1.0}{
\begin{tabular}{crrrrrrrrrr} 
\hline
{\bf inst.} & \multicolumn{2}{c}{\textsc{Heuristic}} &$\;$& \multicolumn{3}{c}{\textsc{Ms-Heur}} &$\;$& \multicolumn{3}{c}{\textsc{Hyb-Ea}} \\ \cline{2-3} \cline{5-7} \cline{9-11}
               & \textbf{result} & \textbf{time} && \textbf{best} & \textbf{avg.} & \textbf{time} && \textbf{best} & \textbf{avg.} & \textbf{time} \\ \hline
Real\_1        & 238 & $<1$ && 181     & 178.30  & 22.34 	&& \ccg247	  & 237.70  & 22.20 \\
Real\_2        & 260 & $<1$ && 206     & 199.10  & 17.27	&& \ccg285	  & 283.40  & 23.12 \\
Real\_3        & 265 & $<1$ && 222     & 218.80  & 16.67	&& \ccg288	  & 280.30  & 26.79 \\
Real\_4        & \ccg373 & $<1$ && 298     & 295.00  & 16.51	&& 369	  & 369.00  & 7.59 \\
Real\_5        & 152 & $<1$ && 133     & 131.10  & 13.87	&& \ccg175	  & 174.30  & 14.96 \\
Real\_6        & 183 & $<1$ && 165     & 160.80  & 9.75		&& \ccg208	  & 203.90  & 20.42 \\
Real\_7        & 316 & $<1$ && 248     & 240.90  & 12.12	&& \ccg328	  & 326.60  & 17.89 \\
Real\_8        & 153 & $<1$ && 141     & 138.50  & 11.78	&& \ccg174	  & 170.00  & 24.82 \\
Real\_9        & 285 & $<1$ && 231     & 223.60  & 13.05	&& \ccg299	  & 297.50  & 22.04 \\
Real\_10       & 343 & $<1$ && 276     & 267.60  & 16.24	&& \ccg352	  & 351.50  & 12.05 \\ \hline
\end{tabular}}
\end{table*}

\section{Conclusion}
\label{sec:conclusions}

In this paper we proposed a simple, hybrid evolutionary algorithm for solving the so-called longest arc-preserving common subsequence problem. The most important feature of this algorithm is a crossover component based on solution merging. At each iteration, the best solution found so far is merged with randomly generated solutions, and a general purpose integer linear programming solver is used to find the best solution within the resulting set of assignments. The results show that the algorithm is superior to the only existing heuristic from the literature. Moreover, we have shown that the solution merging component is an essential part of the algorithm. 

In future work we will try to replace the probabilistic way of constructing solutions by a probabilistic version of the Smith and Waterman algorithm. In this way it might be possible to avoid situations such as the one for real-life instance Real\_4, where our algorithm was not able to outperform the existing heuristic.


\section*{Acknowledgment}

This work was funded by project TIN2012-37930-C02-02 (Spanish Ministry for Economy and Competitiveness, FEDER funds from the European Union) and project SGR 2014-1034 (AGAUR, Generalitat de Catalunya). Our experiments have been executed in the High Performance Computing environment managed by the {\sc rd}lab at the Technical University of Barcelona (\texttt{http://rdlab.cs.upc.edu}) and we would like to thank them for their support.


\end{document}